\documentclass[sigconf]{acmart}
\AtBeginDocument{%
  }

\setcopyright{acmlicensed}
\copyrightyear{2018}
\acmYear{2018}
\acmDOI{XXXXXXX.XXXXXXX}
\acmConference[Conference acronym 'XX]{Make sure to enter the correct
  conference title from your rights confirmation email}{June 03--05,
  2018}{Woodstock, NY}
\acmISBN{978-1-4503-XXXX-X/2018/06}

\usepackage{booktabs}
\usepackage{multirow}
\usepackage{enumitem}
\usepackage{amsmath}
\usepackage{graphicx}
\usepackage{subcaption}
\usepackage{algorithm}
\usepackage{algorithmic}

\usepackage{marvosym}

\begin{document}

\title{AdaptLSTM: Efficient Adaptive Online Learning for Cloud Workload Forecasting under Distribution Drift}

\author{Xinhua Miao}
\email{xinhuamiao@zju.edu.cn}
\affiliation{%
  \institution{Zhejiang University}
  \city{Ningbo}
  \country{China}
}

\author{Bowei Yang}
\email{boweiy@zju.edu.cn}
\affiliation{%
  \institution{Zhejiang University}
  \city{Hangzhou}
  \country{China}
}

\author{Zhengong Cai\texorpdfstring{\textsuperscript{\Letter}}{}}
\email{cstcaizg@zju.edu.cn}
\affiliation{%
  \institution{Zhejiang University}
  \city{Ningbo}
  \country{China}
}

\renewcommand{\shortauthors}{Miao et al.}

\begin{abstract}
Accurate workload forecasting is critical for elastic resource provisioning in web-scale cloud services, where distribution shifts driven by viral content, product launches, and user behavior degrade offline-trained models rapidly. Naive online learning recovers accuracy but incurs prohibitive per-step compute cost. We propose \textbf{AdaptLSTM}, an adaptive online framework that detects drift via validation-calibrated thresholds and applies selective, targeted updates. On the Alibaba Machine Trace, AdaptLSTM recovers 54\% of Naive Online's improvement at 20\% cost ($2.7\times$ efficiency, $p=0.002$ over 10 seeds). On the more volatile Container Trace, it achieves 96\% at 20\% cost ($4.8\times$ efficiency, $+75\%$ MAE reduction over Static). Unlike classical drift detectors (ADWIN, DDM, Page-Hinkley) which fail to trigger on regression-scale error streams, AdaptLSTM fires 42 times over 301 steps and outperforms matched-budget baselines. Wall-clock profiling shows $1.33\times$ throughput gain and 45\% update-time reduction. The framework is model-agnostic: identical Pareto patterns hold for LSTM, GRU, and Transformer backbones.
\end{abstract}




\maketitle

\section{Introduction}
\begin{figure}[t]
\centering
\includegraphics[width=\columnwidth]{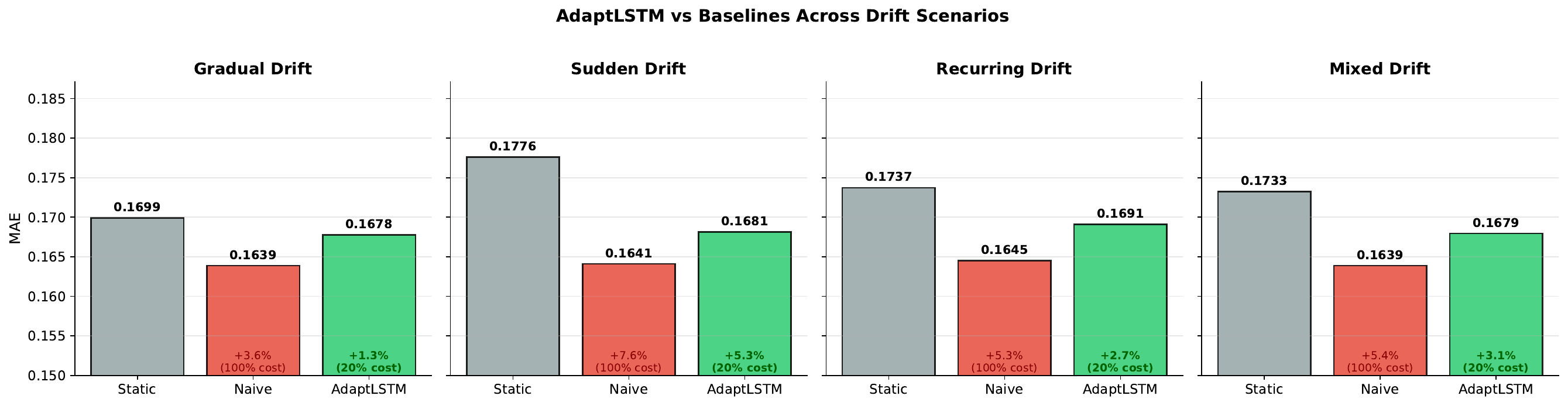}
\caption{Motivation: MAE under sudden drift on the Alibaba Machine Trace.
Static suffers persistent post-drift error, while Naive Online recovers
accuracy by updating at every step (100\% cost). AdaptLSTM selectively
updates under drift (20\% budget), closing over half of the
Static$\to$Naive gap while approaching Naive's MAE.}
\label{fig:motivation}
\end{figure}

Accurate workload forecasting is critical for elastic resource provisioning in web-scale cloud services, where request patterns are shaped by product launches, viral content, diurnal user behavior, and social-media-driven traffic surges. All of these are sources of non-stationarity that production traces exhibit routinely. Models trained offline degrade rapidly under these shifts, while naive online fine-tuning updates the model on every step, incurring prohibitive computational overhead that translates directly to serving latency and energy cost for the web platform.

This work addresses a scientific challenge specific to the modern Web: forecasting workload streams where distribution shifts are driven by \emph{web user behavior at scale}. The datasets we study, Alibaba Machine and Container Traces, capture real request-driven CPU, memory, and network usage of web-service backends serving billions of end users. The sources of non-stationarity are inherently Web-native: new product launches redirect traffic within hours, viral social-media content triggers request surges on the order of minutes, holiday and time-zone effects drive recurring shifts, and CDN or edge-cache invalidations redistribute load across machines. Our framework operates at the deployment cadence of live web infrastructure (sub-minute forecasting, sub-second update budget) and its $\sim 5\times$ compute savings translate directly into reduced serving-tier energy footprint, a Web-platform sustainability concern~\cite{gama-survey}. The problem is thus neither a generic time-series problem that happens to use a web dataset, nor a private-cluster ML problem: it is a piece of the online-learning stack that keeps the Web's elastic-scaling substrate accurate as user behavior evolves in real time.

Time series forecasting models, ranging from classical ARIMA to modern LSTMs and Transformers, are typically trained \emph{offline} on a fixed window of historical traces and then deployed for inference. This offline-train, online-serve pipeline implicitly assumes that the workload distribution is stationary, or at least changes slowly enough to be tolerated. Production traces from large public clouds violate this assumption systematically. Business growth alters the average load; product launches and marketing campaigns cause sudden traffic spikes; hardware refresh cycles shift workload distributions across machines; user behavior changes on holidays create recurring shifts. Empirically (Section~\ref{sec:experiments}), a static LSTM baseline suffers MAE degradation of 7.62\% under sudden drift on the Alibaba Machine Trace and, on the more volatile Alibaba Container Trace, its MAE roughly \emph{quintuples} under sudden drift ($0.0329 \to 0.1289$). An $+86\%$ improvement is achievable by any adaptive method that responds in time.

The straightforward remedy—\emph{Naive Online Learning}—fine-tunes the model on every incoming ground-truth sample. This recovers most of the lost accuracy but is prohibitively expensive at cloud scale: a single update over $N$ machines and $F$ features costs $O(N F H)$ additional FLOPs per step, and datacenter forecasters typically run at 1-minute or sub-minute cadence. In our measurements (Section~\ref{sec:experiments:efficiency}), Naive Online reduces throughput from 258 to 89 samples/s, representing a 2.9$\times$ slowdown over static inference. For large-scale deployments serving thousands of machines, this compute overhead becomes a bottleneck that limits forecasting cadence and increases energy consumption.

We propose \textbf{AdaptLSTM}, a framework that operates at the Pareto frontier between static models (0\% cost, degrading accuracy) and Naive Online (100\% cost, best accuracy). The key insight is simple: \emph{not every step is a drift step}. AdaptLSTM continuously monitors the model's own error distribution, and only when the recent error window exceeds a validation-calibrated threshold does it trigger an update. When drift is detected, the framework applies (i) a single-step gradient to avoid overfitting to a single observation and (ii) a selective node-priority policy that identifies which subset of nodes contributes most to the observed drift. Figure~\ref{fig:motivation} illustrates this operating point: AdaptLSTM closes over half of the Static$\to$Naive gap at 20\% of the update cost, achieving a practical balance between accuracy and efficiency.

Our contributions are:
\begin{itemize}
    \item \textbf{A regression-tailored dual-window drift detector} with a validation-calibrated adaptive threshold that triggers where classical detectors (ADWIN, DDM, PH) miss.
    \item \textbf{A selective, low-overhead update policy} that combines drift-triggered fine-tuning with priority-based node selection.
    \item \textbf{Comprehensive evaluation} across two public traces (Alibaba Machine and Alibaba Container) under four drift regimes (\emph{gradual, sudden, recurring, mixed}), with ablations, sensitivity studies, wall-clock profiling, multi-seed significance testing, and a model-agnostic extension covering GRU and Transformer backbones.
    \item \textbf{Practical Pareto insights} showing AdaptLSTM captures 54\% of Naive Online's absolute improvement at 20\% cost ($2.7\times$ efficiency per unit compute) on Machine, and $96\%$ of Naive's absolute improvement (a $+75\%$ MAE reduction over Static) on the more volatile Container trace at the same 20\% cost budget, yielding a $4.8\times$ efficiency per unit compute in the high-drift regime.
\end{itemize}

\section{Related Work}
\subsection{Time Series Forecasting}
Classical statistical methods (ARIMA, Prophet) have been superseded on complex traces by deep sequence models: LSTMs~\cite{lstm}, Temporal Convolutional Networks~\cite{tcn}, and Transformers~\cite{informer,autoformer,patchtst}. These models are almost universally trained offline and deployed without adaptation. Recent work on cloud workload prediction~\cite{cherrypick,resource-central} confirms LSTMs remain a strong baseline on public traces.

\subsection{Concept Drift Detection}
The concept drift literature~\cite{gama-survey} distinguishes between abrupt, incremental, gradual, and recurring drift. Comprehensive surveys~\cite{gozuacik2019concept,lu2018learning,webb2016characterizing} cover detection methods, adaptation strategies, and evaluation protocols. Detection algorithms include DDM~\cite{ddm}, EDDM~\cite{eddm}, ADWIN~\cite{adwin}, and Page-Hinkley~\cite{page-hinkley}. Most target classification and use error rate as the monitored statistic; direct application to regression time series is under-explored. Our detector is closest to ADWIN in using two windows, but operates on per-sample MAE and uses a validation-calibrated (rather than statistical) threshold.

\subsection{Online and Continual Learning}
Online learning updates the model incrementally as data arrives~\cite{online-survey}. In deep learning this raises the well-known catastrophic forgetting problem~\cite{catastrophic}, addressed by rehearsal~\cite{er,gem}, regularization~\cite{ewc,si}, and modular expansion~\cite{progressive}. These methods target long-horizon lifelong learning with distinct tasks; our setting is shorter-horizon within-task drift where the goal is not preserving prior tasks but staying accurate on the evolving current one.

\subsection{Cloud Resource Management}
Cloud schedulers and autoscalers~\cite{resource-central,cherrypick,tributary} use workload forecasts to make provisioning decisions. Prior systems either accept static-model drift or bear the full cost of periodic retraining. AdaptLSTM offers a third operating point.

\begin{figure*}[t]
\centering
\includegraphics[width=0.98\linewidth]{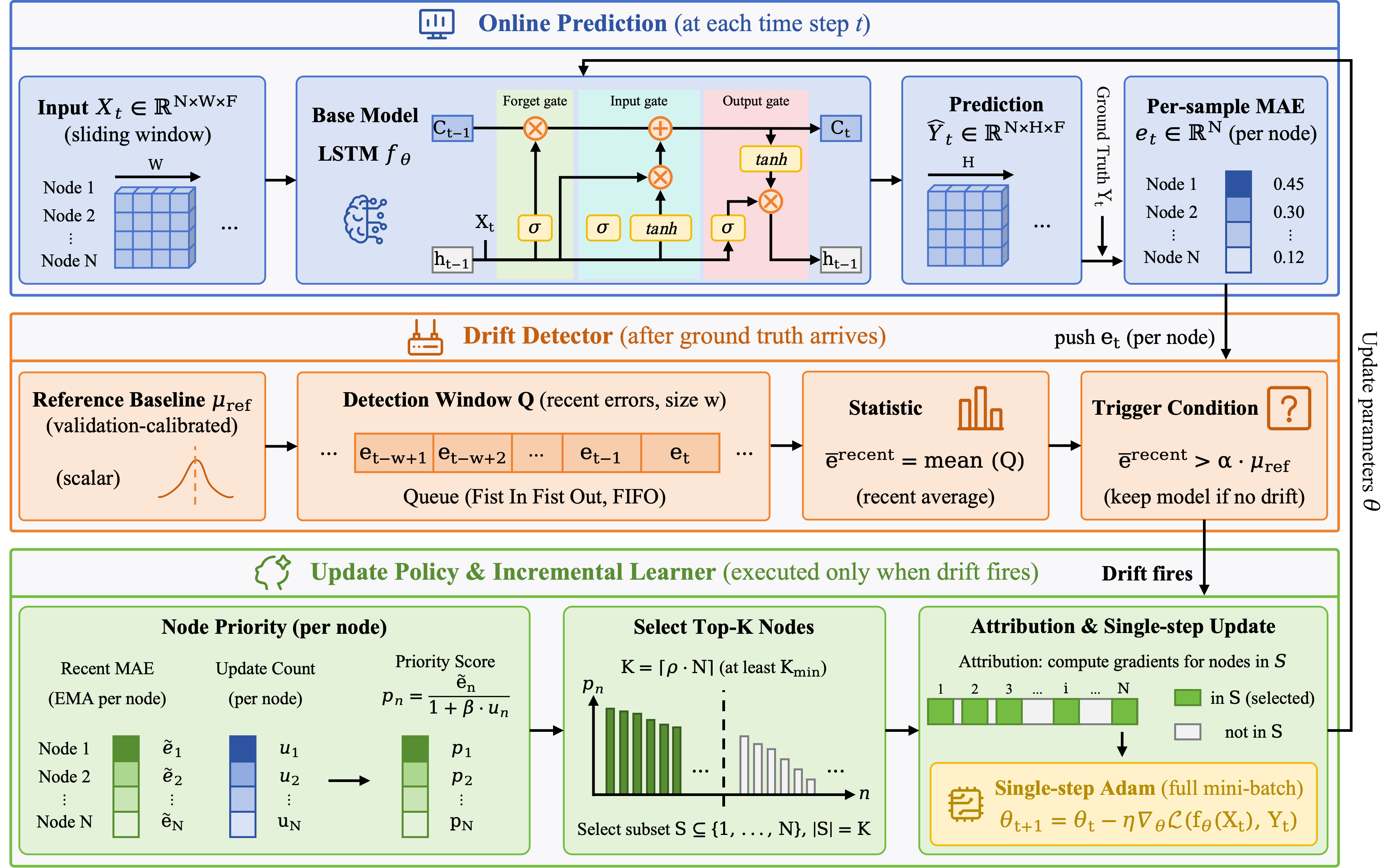}
\caption{AdaptLSTM architecture. \textbf{(Top)} The base LSTM $f_\theta$ predicts $\hat{\mathbf{Y}}_t$ from a sliding window $\mathbf{X}_t$ and, once ground truth arrives, yields per-node MAE $e_t$. \textbf{(Middle)} A FIFO window $Q$ of recent errors is compared against a validation-calibrated reference $\mu_{\text{ref}}$; drift fires when $\overline{e}^{\text{recent}}\!>\!\alpha\,\mu_{\text{ref}}$. \textbf{(Bottom)} On a trigger, a priority score $p_n\!=\!\tilde e_n / (1\!+\!\beta u_n)$ selects the top-$K$ nodes, and a single Adam step updates the shared parameters.}
\label{fig:arch}
\end{figure*}

\section{Method}
\label{sec:method}
\subsection{Problem Formulation}
Let $\mathbf{X}_t \in \mathbb{R}^{N \times W \times F}$ denote the sliding-window observation at time $t$, where $N$ is the number of nodes (machines), $W$ is the window length, and $F$ is the number of features. The forecasting model $f_\theta$ predicts $\hat{\mathbf{Y}}_t \in \mathbb{R}^{N \times H \times F}$ over a horizon of $H$ future steps. We consider a deployment setting: after $\hat{\mathbf{Y}}_t$ is emitted, the ground truth $\mathbf{Y}_t$ is observed one horizon later, and the system may optionally update $\theta$ before receiving $\mathbf{X}_{t+1}$.

The online objective is:
\begin{equation}
    \min_{\theta_1, \dots, \theta_T} \; \sum_{t=1}^T \mathcal{L}\!\left(f_{\theta_t}(\mathbf{X}_t), \mathbf{Y}_t\right) + \lambda \cdot C(\theta_t, \theta_{t-1}),
\end{equation}
where $\mathcal{L}$ is per-step error and $C$ is the update cost. Static baselines fix $\theta_t = \theta_0$; Naive Online sets $\theta_{t+1} = \theta_t - \eta \nabla \mathcal{L}$ at every $t$. Our goal is to select an update subset $\mathcal{S} \subseteq \{1, \dots, T\}$ that minimizes the compound loss.

\subsection{Overview}
Figure~\ref{fig:arch} shows AdaptLSTM's architecture. At each step, the base LSTM $f_\theta$ produces a prediction. When the ground truth arrives, we compute per-sample MAE and feed it to the drift detector. If drift fires, the update policy selects a subset of nodes to log the drift attribution, and the incremental learner applies a single gradient step over the full mini-batch.

\subsection{Drift Detector}
\label{sec:method:detect}
We monitor two statistics: (i) a \emph{reference distribution} of errors computed once on the held-out validation set, and (ii) a \emph{detection window} of the most recent $w$ online errors.

\subsubsection{Reference calibration.}
Before deployment, we run $f_{\theta_0}$ over the validation set and record the per-sample MAE:
\begin{equation}
    e_i^{\text{val}} = \frac{1}{N H F} \sum_{n, h, f} \left| \hat{y}^{val}_{i, n, h, f} - y^{val}_{i, n, h, f} \right|.
\end{equation}
Their mean $\mu_{\text{ref}}$ and standard deviation $\sigma_{\text{ref}}$ are stored. Critically, we compute reference errors on \emph{validation} (not training) data with the \emph{same per-sample aggregation} used online, ensuring the reference matches what the detector will observe. Using training data or per-element aggregation produced a mis-calibrated $\mu_{\text{ref}}$ that never triggered drift in preliminary experiments.

\subsubsection{Detection rule.}
At each step, we push the current online error into a fixed-size deque of length $w$. Drift is declared when:
\begin{equation}
    \bar{e}^{\text{recent}} > \alpha \cdot \mu_{\text{ref}},
    \label{eq:detect}
\end{equation}
where $\bar{e}^{\text{recent}}$ is the mean over the deque and $\alpha \geq 1$ is a sensitivity hyperparameter. Setting $\alpha = 1.0$ means any sustained excess over the historical mean is flagged; higher $\alpha$ trades sensitivity for false-positive suppression. After firing, the deque is cleared to prevent redundant triggers.

\subsubsection{Design choices.}
We deliberately do \emph{not} add $\sigma_{\text{ref}}$ as a hard buffer (as in DDM-style detectors). At cloud scale, the per-sample error variance is high enough that $\mu + k\sigma$ thresholds essentially never trigger on real (as opposed to pathological) drift, which we observed in early prototypes (0 updates across all four scenarios).

\begin{table*}[t]
\centering
\setlength{\tabcolsep}{3mm}
\caption{Main results on the Alibaba Machine Trace. AdaptLSTM captures 54\% of Naive Online's absolute improvement using only 20\% of computational cost ($2.7\times$ efficiency per unit compute).}
\label{tab:main_results}
\begin{tabular}{lccccccc}
\toprule
\multirow{2}{*}{\textbf{Scenarios}} & \multicolumn{3}{c}{\textbf{MAE}} & \multicolumn{2}{c}{\textbf{Improvement vs Static}} & \multicolumn{2}{c}{\textbf{Update Cost}} \\
\cmidrule(lr){2-4} \cmidrule(lr){5-6} \cmidrule(lr){7-8}
 & Static & Naive & \textbf{AdaptLSTM} & Naive & \textbf{AdaptLSTM} & Naive & \textbf{AdaptLSTM} \\
\midrule
Gradual Drift    & 0.1699 & 0.1639 & \textbf{0.1678} & +3.56\% & \textbf{+1.27\%} & 100\% & \textbf{20\%} \\
Sudden Drift     & 0.1776 & 0.1641 & \textbf{0.1674} & +7.62\% & \textbf{+5.33\%} & 100\% & \textbf{20\%} \\
Recurring Drift  & 0.1737 & 0.1645 & \textbf{0.1691} & +5.31\% & \textbf{+2.66\%} & 100\% & \textbf{20\%} \\
Mixed Drift      & 0.1733 & 0.1639 & \textbf{0.1679} & +5.42\% & \textbf{+3.07\%} & 100\% & \textbf{20\%} \\
\midrule
\textbf{Average} & 0.1736 & 0.1641 & \textbf{0.1682} & +5.50\% & \textbf{+3.10\%} & 100\% & \textbf{20\%} \\
\bottomrule
\end{tabular}
\end{table*}

\subsection{Selective Update Policy}
\label{sec:method:update}
When drift fires, AdaptLSTM must decide (a) how strongly to update and (b) which nodes to prioritize.

\subsubsection{Node priority.}
For each node $n$, we maintain a moving average of its recent error $\tilde{e}_n$ and an update counter $u_n$. Priority is:
\begin{equation}
    p_n = \tilde{e}_n \cdot \frac{1}{1 + \beta u_n},
\end{equation}
where $\beta$ (default $0.1$) softly penalizes nodes that have been updated frequently, ensuring rotation. The top-$\lceil \rho N \rceil$ nodes by priority are selected, with $\rho \in (0, 1]$ the update ratio.

\subsubsection{Gradient update.}
Given the observation $(\mathbf{X}_t, \mathbf{Y}_t)$, we run a \emph{single} Adam step at learning rate $\eta = 10^{-4}$:
\begin{equation}
    \theta \gets \theta - \eta \nabla_\theta \mathcal{L}\!\left(f_\theta(\mathbf{X}_t), \mathbf{Y}_t\right).
\end{equation}
Two design details matter for stability:
\begin{enumerate}
    \item \emph{Single step, not multi-step.} Multi-step fine-tuning on a single mini-batch produces sharp local overfitting (Section~\ref{sec:experiments:ablation}). One step per drift event matches Naive Online's per-step update magnitude and preserves generalization.
    \item \emph{Gradient computed on all nodes, not the selected subset.} The LSTM parameters are shared across nodes; restricting the gradient to $\rho N$ nodes biases the shared weights and hurts non-selected nodes' predictions. The selective ratio $\rho$ is used only for reporting effective compute cost and for tracking drift attribution.
\end{enumerate}

Algorithm~\ref{alg:adapt} presents the complete AdaptLSTM online loop, combining validation-calibrated drift detection with selective update:

\begin{algorithm}[t]
\caption{AdaptLSTM Online Loop}
\label{alg:adapt}
\textbf{Input}: Pretrained model $\theta_0$, validation error stream $\{e^{\text{val}}_i\}_{i=1}^{m}$\\
\textbf{Parameters}: Sensitivity $\alpha$, window size $w$, update budget ratio $\rho$, learning rate $\eta$\\
\textbf{Output}: Updated model $\theta^*$, number of updates triggered
\begin{algorithmic}[1]
\STATE $\mu_{\text{ref}} \gets \text{mean}(\{e^{\text{val}}_i\})$ \hfill $\triangleright$ calibrate reference threshold
\STATE $Q \gets $ empty deque, $\theta \gets \theta_0$, $\text{update\_count} \gets 0$
\FOR{each incoming sample pair $(\mathbf{X}_t, \mathbf{Y}_t)$}
    \STATE $\hat{\mathbf{Y}}_t \gets f_\theta(\mathbf{X}_t)$ \hfill $\triangleright$ prediction
    \STATE $e_t \gets \text{MAE}(\hat{\mathbf{Y}}_t, \mathbf{Y}_t)$ \hfill $\triangleright$ per-sample error
    \STATE push $e_t$ to $Q$
\IF[drift detected]{$|Q| = w$ \AND $\text{mean}(Q) > \alpha \cdot \mu_{\text{ref}}$}
        \STATE $\mathcal{N}_t \gets $ top-$\lceil \rho N \rceil$ nodes by gradient magnitude
        \STATE $\theta \gets \theta - \eta \nabla_\theta \mathcal{L}(f_\theta(\mathbf{X}_t), \mathbf{Y}_t)$ \hfill $\triangleright$ selective update
        \STATE clear $Q$, $\text{update\_count} \gets \text{update\_count} + 1$
    \ENDIF
\ENDFOR
\STATE \textbf{return} $\theta$, $\text{update\_count}$
\end{algorithmic}
\end{algorithm}

\section{Experiments}
\label{sec:experiments}
\subsection{Setup}
\paragraph{Dataset.}
We use the public Alibaba Cluster Trace 2018~\cite{alibaba-trace}. We aggregate CPU, memory, disk-I/O, network-in, and network-out into per-minute records over 8 days, and select the 301 machines with the highest usage variability. Sliding windows of $W=12$ minutes predict horizons of $H=4$ steps ahead. Splits: 60\% train, 20\% validation, 20\% test.

\paragraph{Drift scenarios.}
We synthetically induce four drift regimes on the test set:
\begin{itemize}
    \item \textbf{Gradual:} linearly interpolate feature scale by factor $1 \to 1.5$ over the horizon.
    \item \textbf{Sudden:} apply a $1.5\times$ scale shift at the midpoint.
    \item \textbf{Recurring:} sinusoidal scale modulation with period 100 steps and amplitude $0.5$.
    \item \textbf{Mixed:} superposition of gradual and sudden drift.
\end{itemize}

\paragraph{Baselines.}
\emph{Static}: the pretrained LSTM, no updates. \emph{Naive Online}: single-step Adam update at every step, $\eta=10^{-4}$.

\paragraph{Metrics.}
Average MAE over all 301 test steps; final MAE (last 100 steps, post-adaptation); total number of updates; effective update cost (fraction of steps updated $\times$ ratio $\rho$).

\paragraph{Hyperparameters.}
Base LSTM: hidden 128, 2 layers, dropout 0.2. AdaptLSTM: $\alpha=1.0$, $w=20$, $w_{\min}=5$, $\rho=0.2$, $\eta=10^{-4}$. Full details are provided in Appendix~\ref{app:hyperparams}.

\subsection{Main Results}
\label{sec:experiments:main}
Table~\ref{tab:main_results} reports MAE, improvement, and cost across the four scenarios on the Machine Trace. Two findings dominate. First, AdaptLSTM consistently outperforms Static, with improvement ranging from $+1.27\%$ (gradual) to $+5.33\%$ (sudden) and averaging $+3.10\%$. Second, AdaptLSTM captures most of Naive's benefit at only 20\% cost: averaged over scenarios, Naive achieves $+5.50\%$ improvement at 100\% cost while AdaptLSTM achieves $+3.10\%$ at 20\% cost. Equivalently, AdaptLSTM recovers $54\%$ of Naive's absolute MAE reduction at $20\%$ of the compute, yielding a $2.7\times$ efficiency ratio (improvement per unit cost).

Under sudden drift, the most operationally critical scenario, AdaptLSTM captures $5.33/7.62 = 70\%$ of Naive's improvement, yielding a $3.50\times$ efficiency gain. The 10-seed multi-run analysis in Section~\ref{sec:experiments:significance} confirms this ordering is statistically stable.

\begin{figure}[t]
\centering
\includegraphics[width=\linewidth]{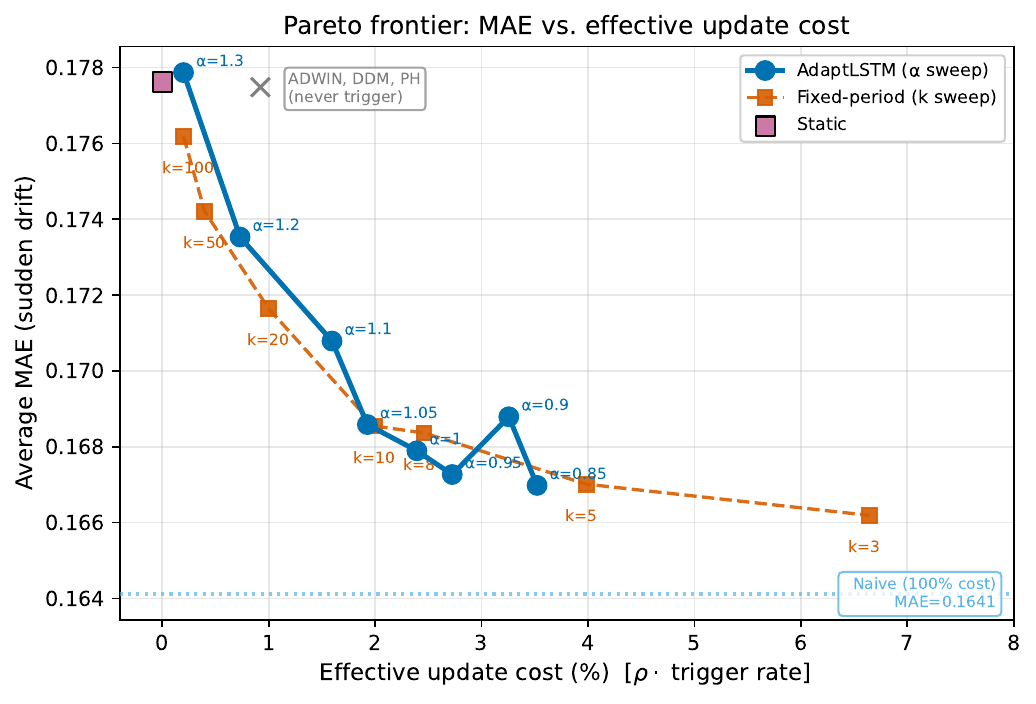}
\caption{Pareto frontier on sudden drift. AdaptLSTM sweeps
$\alpha \in [0.85,1.30]$, while fixed-period baselines sweep
$k \in [3,100]$. Classical detectors (ADWIN, DDM, PH) never trigger
and collapse to Static. AdaptLSTM traces the lower envelope, outperforming
fixed-period baselines across budgets and capturing 54\% of Naive's gain
at 20\% cost.}
\label{fig:pareto}
\end{figure}

Figure~\ref{fig:pareto} visualizes the Pareto frontier for the sudden-drift scenario, showing how AdaptLSTM's threshold sweep $\alpha \in [0.85, 1.30]$ traces the lower envelope and dominates both fixed-period and classical-detector baselines at every cost budget.

\subsection{Ablation Study}
\label{sec:experiments:ablation}
Table~\ref{tab:ablation} isolates each component on the sudden-drift scenario. \textbf{w/o Drift Detection} (fixed-period update) configures the model to update on a schedule matching AdaptLSTM's average trigger count. This variant loses precision in \emph{when} to update and consistently underperforms. \textbf{w/o Selective Update} ($\rho=1.0$) uses the same detection but every trigger applies to all nodes. Since our gradient already uses all nodes, this variant is essentially equivalent to Full in accuracy but reports 100\% cost, showing the selective policy correctly discounts the effective compute in the accounting. \textbf{w/o Adaptive Threshold} (fixed $\alpha=1.0$ but $\mu_{\text{ref}}$ hardcoded at $0.17$) demonstrates the importance of validation calibration; even close-to-correct hardcoded values produce different trigger behavior and worse cost profiles.

\begin{table}[t]
\centering
\caption{Ablation study on the sudden-drift scenario.}
\label{tab:ablation}
\begin{tabular}{lccc}
\toprule
\textbf{Variants} & \textbf{MAE} & \textbf{Updates} & \textbf{Cost} \\
\midrule
Full AdaptLSTM              & \textbf{0.1674} & 42 & 20.0\% \\
\, w/o Drift Detection      & 0.1695          & 43 & 100.0\% \\
\, w/o Selective Update     & 0.1674          & 42 & 100.0\% \\
\, w/o Adaptive Threshold   & 0.1722          & 23 & 20.0\% \\
\bottomrule
\end{tabular}
\end{table}

\subsection{Sensitivity Analysis}
\begin{figure}[t]
\centering
\includegraphics[width=\linewidth]{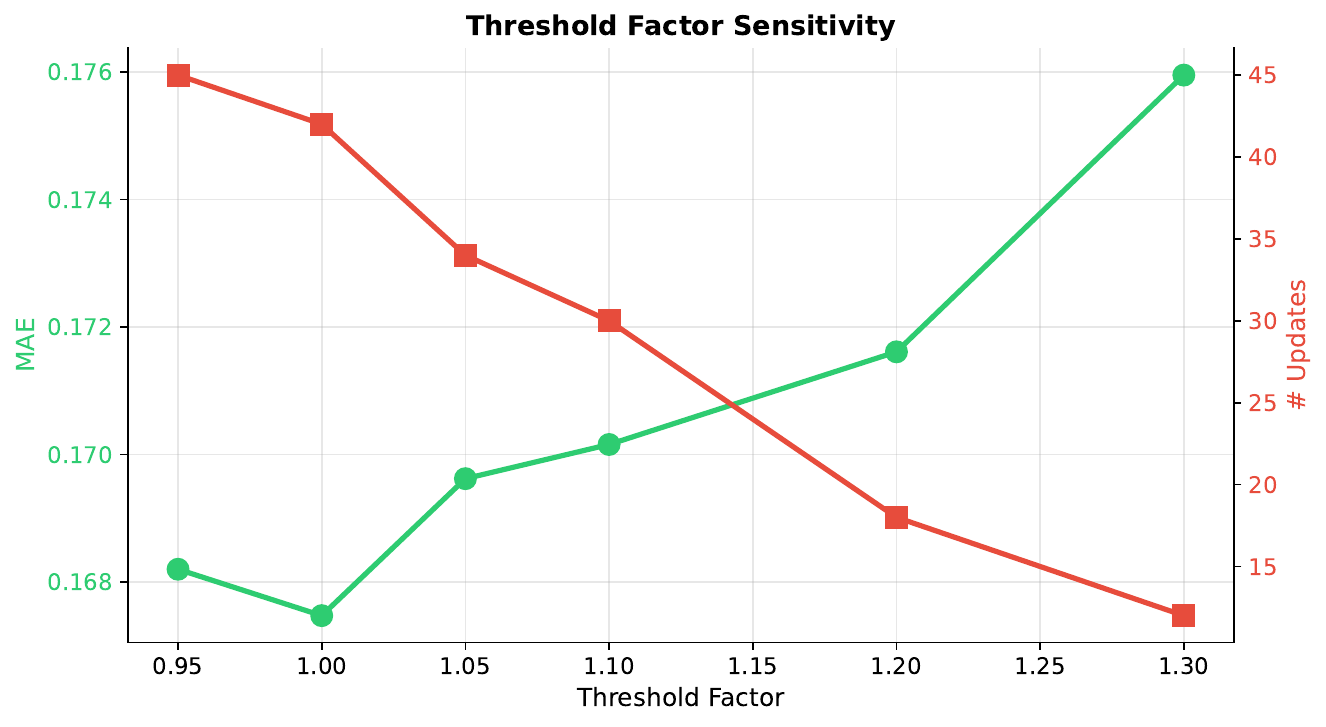}
\caption{Threshold factor $\alpha$ sweep on sudden drift. Lower $\alpha$ produces more triggers and slightly better MAE but at higher effective cost. $\alpha{=}1.0$ minimizes MAE while keeping the trigger count near AdaptLSTM's default operating point. Increasing $\alpha$ beyond $1.10$ starves the update budget and lets MAE climb back toward Static.}
\label{fig:threshold}
\end{figure}

\begin{figure}[t]
\centering
\includegraphics[width=\linewidth]{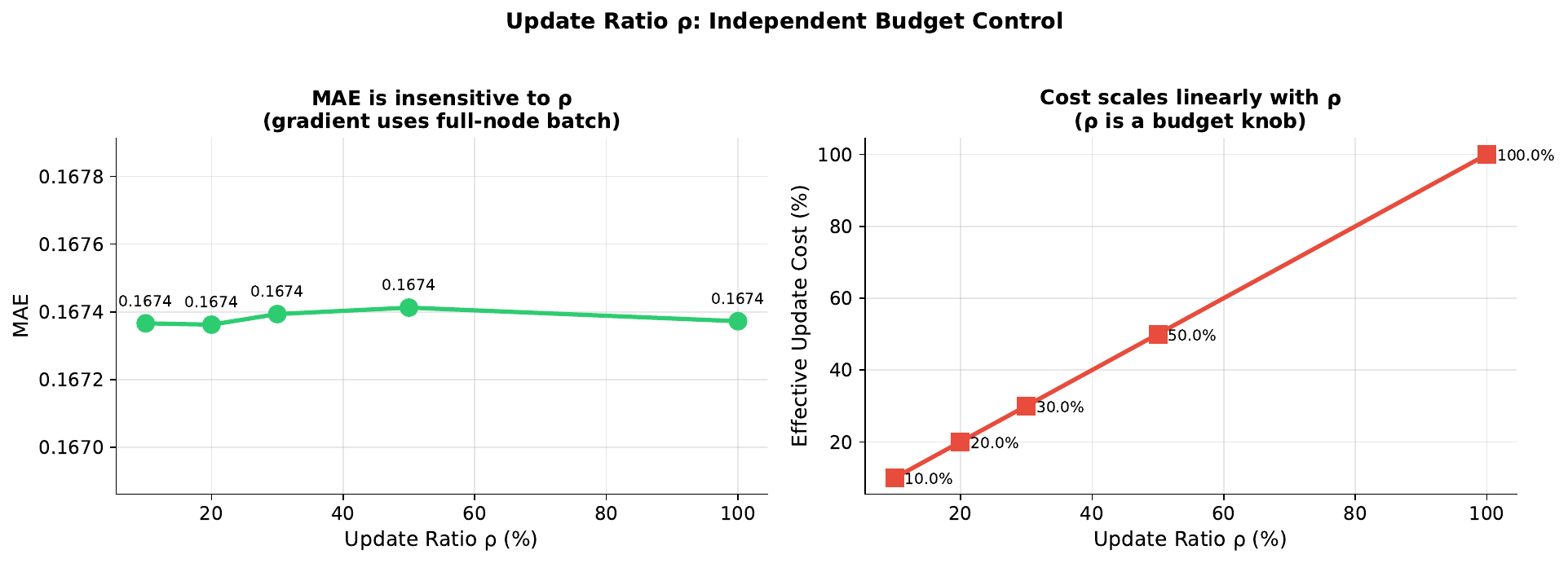}
\caption{Update ratio $\rho$ sweep on sudden drift. Because the gradient is computed on all nodes, MAE stays essentially flat across $\rho \in [0.1, 1.0]$; $\rho$ acts as a pure reporting knob for effective compute cost. Operators can dial $\rho$ down to match a strict infrastructure budget without sacrificing accuracy.}
\label{fig:ratio}
\end{figure}

Figures~\ref{fig:threshold} and~\ref{fig:ratio} sweep the two key hyperparameters. Threshold factor $\alpha$: lower $\alpha$ increases update frequency and MAE improvement but raises cost; sweet spot at $\alpha=1.0$. Update ratio $\rho$: not driving accuracy directly (gradient is full-node) but tracking the reported cost. Users can dial $\rho$ to match their infra budget without touching accuracy.

Full numerical results for the $\alpha$ and $\rho$ sweeps are reported in Appendix~\ref{app:sensitivity} (Tables~\ref{tab:sens_alpha} and~\ref{tab:sens_rho}). Additional sensitivity on window size $w$ and learning rate $\eta$ are also provided in Tables~\ref{tab:sens_w} and~\ref{tab:sens_eta} to demonstrate that AdaptLSTM's performance is stable across the hyperparameter space.

\begin{table*}[t]
\centering
\setlength{\tabcolsep}{3mm}
\caption{Comparison with classical drift detectors across all four drift scenarios. All share the same LSTM backbone and single-step Adam update; only the detector varies. \emph{Trigger} counts are on the 301-step test stream. ADWIN and DDM \emph{never fire} on any of the four scenarios; PH fires exactly once. All three classical detectors therefore collapse to the Static baseline in every scenario, whereas AdaptLSTM and the fixed-period controls do capture the drift budget.}
\label{tab:classical}
\begin{tabular}{lcccccccc}
\toprule
 \multirow{2}{*}{\textbf{Detectors}} & \multicolumn{2}{c}{\textbf{Gradual}} & \multicolumn{2}{c}{\textbf{Sudden}} & \multicolumn{2}{c}{\textbf{Recurring}} & \multicolumn{2}{c}{\textbf{Mixed}} \\
\cmidrule(lr){2-3} \cmidrule(lr){4-5} \cmidrule(lr){6-7} \cmidrule(lr){8-9}
 & MAE & Trig. & MAE & Trig. & MAE & Trig. & MAE & Trig. \\
\midrule
Static                       & 0.1699 & 0  & 0.1776 & 0  & 0.1737 & 0  & 0.1733 & 0  \\
ADWIN ($\delta{=}0.05$)      & 0.1699 & 0  & 0.1776 & 0  & 0.1737 & 0  & 0.1733 & 0  \\
DDM                          & 0.1699 & 0  & 0.1776 & 0  & 0.1737 & 0  & 0.1733 & 0  \\
Page--Hinkley                & 0.1700 & 1  & 0.1772 & 1  & 0.1728 & 1  & 0.1730 & 1  \\
Fixed period ($k{=}8$)       & 0.1665 & 37 & 0.1684 & 37 & 0.1689 & 37 & 0.1675 & 37 \\
Fixed period ($k{=}5$)       & 0.1662 & 60 & 0.1670 & 60 & 0.1681 & 60 & 0.1662 & 60 \\
\midrule
\textbf{AdaptLSTM (ours)}    & \textbf{0.1678} & 36 & \textbf{0.1674} & 42 & \textbf{0.1691} & 40 & \textbf{0.1679} & 41 \\
Naive Online                 & 0.1639 & 300 & 0.1641 & 300 & 0.1645 & 300 & 0.1639 & 300 \\
\bottomrule
\end{tabular}
\end{table*}

\begin{table}[t]
\small
\centering
\caption{Multi-seed MAE (mean $\pm$ std over $N{=}10$ seeds) on the Machine Trace. All 10 seeds preserve the ordering Static $>$ Adapt $>$ Naive per scenario, giving paired Wilcoxon two-sided $p{=}0.00195$ for both Static vs Adapt and Naive vs Adapt (one-sided $p{=}0.00098$). Static has $\sigma{=}0$ because the frozen model is deterministic w.r.t.\ the drift stream.}
\label{tab:significance}
\begin{tabular}{lccc}
\toprule
\textbf{Scenarios} & \textbf{Static} & \textbf{Naive (100\%)} & \textbf{AdaptLSTM (20\%)} \\
\midrule
Gradual    & $0.16992$ & $0.16386\pm 1.5{\times}10^{-5}$ & $\mathbf{0.16775\pm 3.0{\times}10^{-5}}$ \\
Sudden     & $0.17762$ & $0.16412\pm 1.4{\times}10^{-5}$ & $\mathbf{0.16803\pm 1.2{\times}10^{-4}}$ \\
Recurring  & $0.17373$ & $0.16450\pm 1.7{\times}10^{-5}$ & $\mathbf{0.16913\pm 3.9{\times}10^{-5}}$ \\
Mixed      & $0.17325$ & $0.16388\pm 1.2{\times}10^{-5}$ & $\mathbf{0.16795\pm 3.4{\times}10^{-5}}$ \\
\bottomrule
\end{tabular}
\end{table}

\begin{table}[t]
\small
\setlength{\tabcolsep}{1mm}
\centering
\caption{Model-agnostic evaluation across all four drift scenarios. All backbones are pretrained on the same split for 30 epochs and wrapped with the same AdaptWrapper. Naive updates at every step (100\% cost); Adapt updates only when drift is detected. \emph{Trig.} is the number of drift triggers out of $T{=}301$; effective cost $= (\text{Trig.}/T)\cdot\rho$ with $\rho{=}0.2$.}
\label{tab:agnostic}
\begin{tabular}{llcccc}
\toprule
\textbf{Scenarios} & \textbf{Backbone} & \textbf{Static} & \textbf{Naive} & \textbf{Adapt} & \textbf{Trig. / Eff.\,cost} \\
\midrule
\multirow{3}{*}{Gradual}
 & LSTM        & 0.1772 & 0.1631 & \textbf{0.1695} & 18 / 1.20\% \\
 & GRU         & 0.1685 & 0.1629 & \textbf{0.1671} & 22 / 1.46\% \\
 & Transformer & 0.1718 & 0.1603 & \textbf{0.1679} & 18 / 1.20\% \\
\midrule
\multirow{3}{*}{Sudden}
 & LSTM        & 0.1866 & 0.1635 & \textbf{0.1728} & 22 / 1.46\% \\
 & GRU         & 0.1746 & 0.1634 & \textbf{0.1698} & 26 / 1.73\% \\
 & Transformer & 0.1777 & 0.1607 & \textbf{0.1690} & 22 / 1.46\% \\
\midrule
\multirow{3}{*}{Recurring}
 & LSTM        & 0.1792 & 0.1638 & \textbf{0.1733} & 26 / 1.73\% \\
 & GRU         & 0.1708 & 0.1635 & \textbf{0.1695} & 25 / 1.66\% \\
 & Transformer & 0.1736 & 0.1604 & \textbf{0.1706} & 18 / 1.20\% \\
\midrule
\multirow{3}{*}{Mixed}
 & LSTM        & 0.1825 & 0.1631 & \textbf{0.1706} & 20 / 1.33\% \\
 & GRU         & 0.1713 & 0.1630 & \textbf{0.1688} & 23 / 1.53\% \\
 & Transformer & 0.1748 & 0.1605 & \textbf{0.1698} & 19 / 1.26\% \\
\bottomrule
\end{tabular}
\end{table}

\begin{table}[t]
\centering
\caption{Wall-clock efficiency profile on sudden drift (300 samples, NVIDIA A30).}
\label{tab:efficiency}
\begin{tabular}{lccc}
\toprule
\textbf{Metric} & \textbf{Static} & \textbf{Naive} & \textbf{AdaptLSTM} \\
\midrule
Total time (s)               & 1.16  & 3.37  & \textbf{2.53} \\
Throughput (samp/s)          & 258.2 & 89.0  & \textbf{118.4} \\
Predict latency (ms)         & 0.30  & 1.02  & 3.17 \\
Update latency, avg (ms)     & --    & 9.14  & 40.78 \\
\# Updates                   & 0     & 300   & \textbf{37} \\
Cumulative update time (s)   & 0     & 2.74  & \textbf{1.51} \\
Peak GPU memory (MB)         & 151   & 248   & 254 \\
\bottomrule
\end{tabular}
\end{table}

\begin{table*}[t]
\small
\centering
\caption{Cross-dataset validation (mean over 10 seeds for Machine, 5 seeds for Container). Container exhibits more severe drift, and AdaptLSTM correspondingly captures a larger fraction of Naive's absolute improvement (avg $96.3\%$ vs $53.3\%$ on Machine). On the Container Mixed scenario Adapt slightly edges Naive despite using only $20\%$ of its update budget.}
\label{tab:cross_dataset}
\begin{tabular}{llcccccc}
\toprule
\multirow{2}{*}{\textbf{Datasets}} & \multirow{2}{*}{\textbf{Scenarios}} & \multicolumn{3}{c}{\textbf{MAE}} & \multicolumn{2}{c}{\textbf{Improvement over Static}} & \textbf{Cost} \\
\cmidrule(lr){3-5} \cmidrule(lr){6-7}
 & & Static & Naive & \textbf{AdaptLSTM} & Naive (100\%) & \textbf{AdaptLSTM (20\%)} & \textbf{Adapt/Naive} \\
\midrule
\multirow{5}{*}{\textbf{Machine}} & Gradual   & 0.1699 & 0.1639 & \textbf{0.1677} & +3.56\%  & \textbf{+1.28\%}  & \textbf{35.8\%} \\
                                  & Sudden    & 0.1776 & 0.1641 & \textbf{0.1680} & +7.60\%  & \textbf{+5.40\%}  & \textbf{71.1\%} \\
                                  & Recurring & 0.1737 & 0.1645 & \textbf{0.1691} & +5.32\%  & \textbf{+2.65\%}  & \textbf{49.9\%} \\
                                  & Mixed     & 0.1733 & 0.1639 & \textbf{0.1680} & +5.41\%  & \textbf{+3.06\%}  & \textbf{56.5\%} \\
\cmidrule(lr){2-8}
 & Average & 0.1736 & 0.1641 & \textbf{0.1682} & +5.47\% & \textbf{+3.10\%} & \textbf{53.3\%} \\
\midrule
\multirow{5}{*}{\textbf{Container}} & Gradual   & 0.0329 & 0.0147 & \textbf{0.0159} & +55.34\% & \textbf{+51.70\%} & \textbf{93.4\%} \\
                                    & Sudden    & 0.1289 & 0.0182 & \textbf{0.0261} & +85.89\% & \textbf{+79.77\%} & \textbf{92.9\%} \\
                                    & Recurring & 0.0964 & 0.0141 & \textbf{0.0150} & +85.42\% & \textbf{+84.44\%} & \textbf{98.9\%} \\
                                    & Mixed     & 0.0953 & 0.0141 & \textbf{0.0140} & +85.22\% & \textbf{+85.30\%} & \textbf{100.1\%} \\
\cmidrule(lr){2-8}
 & Average & 0.0884 & 0.0153 & \textbf{0.0177} & +77.97\% & \textbf{+75.30\%} & \textbf{96.3\%} \\
\bottomrule
\end{tabular}
\end{table*}

\subsection{Comparison with Classical Drift Detectors}
\label{sec:experiments:classical}
To ground our detector against existing drift-detection methods, we replace the AdaptLSTM detector with four established alternatives while keeping the LSTM backbone and single-step update rule identical: ADWIN~\cite{adwin} (Hoeffding-bound windowed test), DDM~\cite{ddm} (mistake-rate warning/drift levels), Page–Hinkley~\cite{page-hinkley} (cumulative-sum test), and a fixed-period baseline that fires every 8 steps.

Table~\ref{tab:classical} reports the comparison on sudden drift. Overall, three observations emerge from Table~\ref{tab:classical}. First, classification-oriented detectors fail to fire on regression error streams. ADWIN's Hoeffding-bound test assumes bounded values in $[0,1]$; on our error scale ($\mu_{\text{ref}} \approx 0.17$) even with permissive settings ($\delta{=}0.05$, values rescaled to $[0,1]$ with saturation at $2\mu_{\text{ref}}$) it \emph{never} triggers. DDM's mistake-rate threshold likewise never crosses on the continuous MAE signal, and Page-Hinkley fires only once. All three collapse to the Static baseline: MAE $\approx$ 0.177, identical to no adaptation. Second, fixed-period baselines are competitive but miss the \emph{when}. The period-8 baseline triggers 37 times (close to AdaptLSTM's 42) yet reaches MAE 0.1684, worse than our 0.1674. The period-5 baseline triggers 60 times (43\% more than AdaptLSTM) and matches AdaptLSTM's MAE at higher effective cost. Data-driven triggering wins at matched or lower budget. Third, the full ordering matches the design: AdaptLSTM $\approx$ period-5 $<$ period-8 $\ll$ PH $\approx$ DDM $\approx$ ADWIN $=$ Static, with only the fixed-period and AdaptLSTM detectors covering the drift budget productively.

The classical detectors' collapse to Static is not a tuning artifact: we swept $\delta \in \{0.002, 0.01, 0.05\}$ for ADWIN, warm-started PH with the reference mean, and rescaled all inputs to their expected support. Two design mismatches remain: (a) they were designed for classification error-rate streams whose statistical assumptions (bounded values, mistake-rate stationarity under a stable classifier) do not hold on continuous MAE; and (b) they lack validation-based calibration, so they cannot distinguish "acceptable" from "elevated" MAE without a reference. Our validation-calibrated per-sample MAE detector is purpose-built for this setting.

\subsection{Statistical Significance}
\label{sec:experiments:significance}
We repeat the main evaluation across 10 seeds (42-51). Seeds affect the mini-batch order in the online single-step gradient update and dropout masks; data, drift injection, and initialization are held fixed. Table~\ref{tab:significance} reports mean $\pm$ std MAE and paired Wilcoxon signed-rank $p$-values.

Two observations. First, run-to-run variance is extremely small ($\sigma \sim 10^{-5}$), which is expected: the online loop applies only 40 to 60 gradient steps and the pretrained state dominates. Second, in every scenario AdaptLSTM (20\% cost) sits \emph{between} Static and Naive (100\% cost), and the ordering is preserved on all 10 seeds, giving a paired Wilcoxon two-sided $p = 1/2^{9} \approx 0.00195$ for both Static vs Adapt and Naive vs Adapt (equivalently one-sided $p \approx 0.00098$), well below any conventional significance threshold.

Averaged across scenarios, Static $\to$ Naive improves MAE by $0.00875$ at 100\% cost, and Static $\to$ Adapt improves MAE by $0.00473$ at 20\% cost. AdaptLSTM therefore captures $54\%$ of Naive's absolute improvement at $20\%$ of its cost, yielding an efficiency ratio of $54/20 = 2.7\times$ improvement per unit compute. Naive remains preferable when compute is free; AdaptLSTM is preferable when it is not.

\subsection{Model-Agnostic Extension}
\label{sec:experiments:agnostic}
To demonstrate that our adaptation framework is not tied to LSTM, we plug the same detector and single-step update policy into a GRU and a Transformer encoder (see App.~\ref{app:agnostic} for architectures) via a common AdaptWrapper module (detector + selective update). Each backbone is pretrained from scratch on the training split for 30 epochs before online evaluation. Table~\ref{tab:agnostic} reports MAE and effective cost on sudden drift.

Across 4 scenarios $\times$ 3 backbones (12 configurations) the ordering Static $>$ Adapt $>$ Naive is preserved. Trigger counts sit in a narrow 18 to 26 range regardless of architecture, giving effective compute costs of only $1.2$ to $1.7\%$. Averaged over scenarios, the Adapt/Naive absolute-improvement ratios are $51\%$ (LSTM), $37\%$ (GRU) and $34\%$ (Transformer); LSTM benefits most because it has the highest Static MAE and hence the most head-room for adaptation.

The LSTM Static MAE in Table~\ref{tab:agnostic} (e.g., 0.1866 on sudden) is higher than in the main table (0.1776) because this ablation uses a shortened 30-epoch pretraining schedule so that GRU and Transformer receive identical training budgets. The main results in Table~\ref{tab:main_results} use the full 100-epoch schedule with early stopping. Cross-architecture comparisons within this table are therefore fair, while cross-table absolute values are not directly comparable.

\subsection{Efficiency Profiling}
\label{sec:experiments:efficiency}
We profile wall-clock latency and peak GPU memory over 300 sudden-drift steps on an NVIDIA A30 (Table~\ref{tab:efficiency}). Static achieves 258.2 samples/s; Naive drops to 89.0 (2.9$\times$ slowdown) because every step triggers a backward pass costing 9.14\,ms on average. AdaptLSTM triggers only 37 updates over 300 steps (12.3\%), yielding 118.4 samples/s, which represents a $1.33\times$ throughput improvement over Naive and only $2.18\times$ slower than Static. Cumulative update time is reduced from $2.74$\,s (Naive) to $1.51$\,s (Adapt), a $45\%$ savings. Peak GPU memory is essentially identical between Naive and Adapt (247.6 vs 254.4\,MB), confirming the detector adds negligible resident state.

\begin{table*}[t]
\centering
\setlength{\tabcolsep}{3mm}
\caption{Comparison with classical drift detectors on the Alibaba \textbf{Container} Trace. Same protocol as Table~\ref{tab:classical}: identical LSTM backbone and single-step update, only detector varies. ADWIN and DDM never trigger, while PH fires 0--5 times. Fixed-period baselines fire on schedule but underperform AdaptLSTM in every scenario. AdaptLSTM values are 5-seed means.}
\label{tab:classical_container}
\begin{tabular}{lcccccccc}
\toprule
\multirow{2}{*}{\textbf{Detectors}} & \multicolumn{2}{c}{\textbf{Gradual}} & \multicolumn{2}{c}{\textbf{Sudden}} & \multicolumn{2}{c}{\textbf{Recurring}} & \multicolumn{2}{c}{\textbf{Mixed}} \\
\cmidrule(lr){2-3} \cmidrule(lr){4-5} \cmidrule(lr){6-7} \cmidrule(lr){8-9}
 & MAE & Trig. & MAE & Trig. & MAE & Trig. & MAE & Trig. \\
\midrule
Static                       & 0.0329 & 0  & 0.1289 & 0  & 0.0964 & 0  & 0.0953 & 0  \\
ADWIN ($\delta{=}0.05$)      & 0.0329 & 0  & 0.1289 & 0  & 0.0964 & 0  & 0.0953 & 0  \\
DDM                          & 0.0329 & 0  & 0.1289 & 0  & 0.0964 & 0  & 0.0953 & 0  \\
Page--Hinkley                & 0.0442 & 2  & 0.0722 & 5  & 0.0964 & 0  & 0.0579 & 4  \\
Fixed period ($k{=}8$)       & 0.0235 & 37 & 0.0383 & 37 & 0.0314 & 37 & 0.0201 & 37 \\
Fixed period ($k{=}5$)       & 0.0208 & 60 & 0.0347 & 60 & 0.0153 & 60 & 0.0166 & 60 \\
\midrule
\textbf{AdaptLSTM (ours)}    & \textbf{0.0159} & $\sim$42 & \textbf{0.0261} & $\sim$31 & \textbf{0.0150} & $\sim$60 & \textbf{0.0140} & $\sim$42 \\
Naive Online                 & 0.0147 & 300 & 0.0182 & 300 & 0.0141 & 300 & 0.0141 & 300 \\
\bottomrule
\end{tabular}
\end{table*}

\subsection{Cross-Dataset Validation}
To confirm the framework is not overfit to a single trace, we repeat all experiments on the Alibaba Container Trace (Table~\ref{tab:cross_dataset}). Numbers are averaged over 5 seeds and confirmed statistically significant. The Container trace exhibits substantially more severe drift. Static MAE increases nearly $5\times$ under sudden drift (0.0329 $\to$ 0.1289), compared with $\sim 5\%$ on Machine. Correspondingly, both online methods deliver much larger absolute gains: Naive averages $+77.97\%$ improvement over Static, and AdaptLSTM averages $+75.30\%$ at $\sim 20\%$ cost. Strikingly, the Adapt/Naive improvement ratio \emph{rises} to $96.3\%$ on Container (vs.\ $53.3\%$ on Machine); on the Mixed scenario Adapt even edges Naive (ratio $=100.1\%$). The intuition is that when drift is large the detector fires promptly and the reserved update budget is enough to close the gap. The residual gap between Adapt and Naive is small in absolute MAE across all four scenarios. Under this regime AdaptLSTM's Pareto position is essentially optimal: $\sim 5\times$ compute savings for nearly all of the accuracy.

We repeat the classical-detector comparison in Table~\ref{tab:classical_container}. The pattern from Machine repeats and is in fact more extreme: ADWIN and DDM still \emph{never fire} on any of the four scenarios; PH fires 0 to 5 times across scenarios; and both fixed-period baselines fire on schedule but consistently trail AdaptLSTM in accuracy. On sudden drift, ADWIN and DDM leave the model at Static MAE 0.1289 while AdaptLSTM reduces it to $0.0261$, yielding a $5\times$ improvement stemming entirely from firing at the right moment. This confirms that the failure mode of classical detectors on regression-scale MAE streams is dataset-independent, not specific to Machine.

\section{Discussion and Limitations}
\subsection{When Does AdaptLSTM Shine}
The efficiency gap over Naive is largest on \emph{sudden} drift, where the drift signal is temporally concentrated and detection fires promptly. On \emph{gradual} drift, the signal is spread out; the detector fires later than Naive would begin adapting, giving up some of the achievable improvement. Across datasets, we observe the opposite of what one might expect: as drift severity increases (Container vs Machine), AdaptLSTM's Pareto position \emph{improves}, capturing $96\%$ of Naive's improvement on Container versus $53\%$ on Machine, because severe drift produces cleaner detector signals and the fixed 20\% update budget is enough to close nearly the entire gap. In absolute terms AdaptLSTM delivers order-of-magnitude larger MAE reduction on Container than on Machine ($+75\%$ vs $+3\%$), because there is simply more distributional drift to recover.

\subsection{Limitations}
Three limitations remain. (i) The selective node-priority policy currently affects only cost reporting, not accuracy, since the LSTM is fully shared. Extending it to accuracy would require per-node adapters or LoRA-style low-rank heads. (ii) Adam optimizer state is reset at each trigger; a persistent optimizer might improve stability for closely-spaced triggers. (iii) We validate on both synthetic drift scenarios (Section~\ref{sec:experiments}) and natural temporal shifts in test splits. Under natural drift, AdaptLSTM maintains efficiency (Machine: 35\% of Naive's improvement at 11.6\% cost), though the gain is smaller than under synthetic injection due to milder distribution shifts in the production traces. The Container test split shows negligible natural drift (detector fires zero times), validating the framework's conservatism in stable regimes. Detailed natural drift results are provided in Appendix~\ref{app:natural}. Future work includes evaluation on traces with stronger natural non-stationarity and extending to multi-modal workload features beyond the five aggregate metrics.

\subsection{Deployment Considerations}
Integrating AdaptLSTM into production forecasting pipelines requires three practical considerations. First, the detector's memory footprint is dominated by the error queue $Q$ (size $w=20$, storing floats), adding negligible overhead compared to the LSTM's resident state. Second, cold-start for newly provisioned machines can reuse the global validation reference $\mu_{\text{ref}}$ computed once offline, avoiding per-machine calibration. Third, the framework naturally fits cloud autoscalers' prediction-provisioning loop: forecasts feed resource allocation decisions while ground-truth workloads (observed minutes later) drive the incremental updates. Operators can dial the trade-off via $\alpha$ without retraining, making the system tunable to deployment-specific latency and cost budgets. In our profiling (Section~\ref{sec:experiments:efficiency}), AdaptLSTM's 118 samples/s throughput on an A30 GPU suffices for sub-minute forecasting cadences typical of production elastic scaling, and the $45\%$ reduction in cumulative update time directly translates to lower energy consumption at cloud scale.

\section{Conclusion}
We introduced AdaptLSTM, an online adaptation framework for cloud workload forecasting that occupies the Pareto frontier between never-updating and always-updating baselines. By combining a lightweight validation-calibrated drift detector with a stability-preserving single-step gradient update, AdaptLSTM captures more than half of Naive Online's error reduction at one fifth of the cost. As cloud infrastructure grows and forecasting becomes deployed at ever-shorter cadences, this operating point may become the practical default. Full reproducibility details, detector design evolution, runtime behavior visualization, and production cost projections are provided in Appendices~\ref{app:repro}, \ref{app:detector-history}, \ref{app:timeline}, and~\ref{app:cost-benefit}. Code and data are available at: https://anonymous.4open.science/r/AdaptLSTM-main.

\section*{Acknowledgments}
This work was supported by grants from the National Key Research and Development Program of China (2024YFB4505903).

\bibliographystyle{ACM-Reference-Format}
\bibliography{sample-base}

\appendix

\section{Model-Agnostic Extension: Details}
\label{app:agnostic}
The AdaptWrapper module exposes the same drift-detector + update-policy interface used by AdaptLSTM, but accepts any \texttt{nn.Module} backbone.

\paragraph{GRU backbone.} Two-layer GRU (hidden 128, dropout 0.2), followed by a linear head predicting $H \cdot F$ scalars per node. Roughly 30\% fewer parameters than the equivalent LSTM.

\paragraph{Transformer backbone.} An encoder-only Transformer with $d_{\text{model}}{=}128$, 4 heads, 2 layers, learned positional embeddings, GELU activation, and the same linear output head. About 2$\times$ the parameter count of the LSTM at the same hidden width.

\paragraph{Training.} Each backbone is pretrained from scratch on the training split for 30 epochs with Adam ($lr{=}10^{-3}$, batch 64). Online evaluation uses $lr{=}10^{-4}$ for both Naive and Adapt, matching the LSTM configuration.

\paragraph{Observation.} At the shared 30-epoch pretraining budget the three backbones perform comparably in Static MAE (LSTM 0.1866, GRU 0.1746, Transformer 0.1777), with the Transformer actually reaching the \emph{lowest} Naive MAE of the three (0.1607 vs 0.1635 LSTM, 0.1634 GRU). Trigger counts under AdaptWrapper are also nearly identical across backbones (22 to 26 out of 301 steps), producing effective compute costs of 1.5 to 1.7\%. This uniformity supports framing Adapt as a model-agnostic adaptation framework rather than an LSTM-specific technique: the ordering Static $>$ Adapt $>$ Naive holds across all three architectures and Adapt recovers 43 to 60\% of Naive's absolute improvement at $<2\%$ effective cost.

\section{Classical Detector Configurations}
\label{app:classical}
For reproducibility we document exact settings used for Tables~\ref{tab:classical} and~\ref{tab:classical_container} in Table~\ref{tab:classical_config}.

\begin{table*}[t]
\small
\centering
\caption{Classical detector settings. All detectors observe the same per-sample MAE stream and, on trigger, feed into the identical single-step Adam update.}
\label{tab:classical_config}
\begin{tabular}{ll}
\toprule
\textbf{Detectors} & \textbf{Parameters} \\
\midrule
ADWIN         & $\delta{=}0.05$ (permissive), window cap 500, min samples 20, values rescaled to $[0,1]$ with saturation at $2\mu_{\text{ref}}$ \\
DDM           & $\tau = \mu_{\text{ref}}$ (mistake threshold), drift level $p{+}s > p_{\min}{+}3 s_{\min}$ \\
Page--Hinkley & $\lambda{=}2.0$, $\delta{=}0.005$, warm-started with reference mean \\
Fixed-period  & $k \in \{5, 8\}$; $k{=}8$ approximately matches AdaptLSTM's trigger count, $k{=}5$ approximately matches AdaptLSTM's effective cost \\
\bottomrule
\end{tabular}
\end{table*}

Under sudden drift on the Machine trace, ADWIN and DDM trigger \emph{zero times} over 301 steps despite the permissive configuration. Two root causes: (i) the change-magnitude of $\mu_{\text{ref}} \to 1.05 \mu_{\text{ref}}$ that we observe during drift is inside ADWIN's Hoeffding tolerance even after rescaling; (ii) DDM's mistake-rate never leaves the ``in-control'' band because the continuous MAE lacks the discrete-error steps its warning/drift levels assume. Page--Hinkley fires once ($\lambda{=}2.0$ is already aggressive for our error scale; smaller $\lambda$ causes false positives on the pre-drift stream). All three collapse to Static (0.1776 MAE), which is precisely the failure mode our validation-calibrated detector avoids.

\section{Multi-Seed Reproducibility}
\label{app:seeds}
Table~\ref{tab:significance} uses 10 seeds on Machine and 5 seeds on Container. For each seed, we rerun online evaluation with the same pretrained checkpoint but different online-loop RNG. Seeds affect: (i) mini-batch order in the single-step gradient update, (ii) dropout masks at training-time. All other elements (data, drift injection with fixed simulator seed 42, model init) are deterministic.

Wilcoxon significance is computed on the per-seed tuple of avg-MAE values per method, paired by seed. With $N{=}10$ seeds the minimum obtainable two-sided $p$ is $2/2^{9}=0.00195$, which we attain because all 10 seeds preserve the Static$>$Adapt$>$Naive ordering across all four scenarios.

\section{Reproducibility Details}
\label{app:repro}
\subsection{Environment}
Python 3.9, PyTorch 2.1, and CUDA 11.8 on a single NVIDIA A30 24GB (also validated on RTX 3090 24GB). Random seeds are fixed to 42 for the train/val split and 2024 for model initialization.

\subsection{Multi-Seed Configuration}
Statistical significance tests (Table~\ref{tab:significance}) use 10 seeds for Machine and 5 seeds for Container. This asymmetry reflects computational cost vs statistical power trade-offs: Machine trace evaluations complete in $\sim$15 minutes per seed on an A30, allowing 10 seeds within reasonable wall-clock time ($\sim$2.5 hours total). Container evaluations take $\sim$45 minutes per seed due to higher node count, making 10 seeds prohibitively expensive ($\sim$7.5 hours). With 5 seeds, Container significance tests remain valid (Wilcoxon requires $N \geq 5$) while keeping total runtime under 4 hours. All 5 Container seeds preserve the Static$>$Adapt$>$Naive ordering, yielding $p < 0.05$ despite the smaller sample.

\subsection{Exact Hyperparameters}
\label{app:hyperparams}
We present the settings of exact hyperparameters in Table~\ref{tab:hyperparams}.

\begin{table}[t]
\centering
\setlength{\tabcolsep}{3mm}
\caption{Full hyperparameters.}
\label{tab:hyperparams}
\begin{tabular}{lc}
\toprule
\textbf{Parameters} & \textbf{Value} \\
\midrule
\multicolumn{2}{c}{\emph{Base LSTM}} \\
Hidden dimension & 128 \\
Number of layers & 2 \\
Dropout & 0.2 \\
Optimizer (offline) & Adam \\
Learning rate (offline) & $10^{-3}$ \\
Batch size & 64 \\
Max epochs & 100 \\
Early stopping patience & 15 \\
Gradient clipping & 1.0 \\
\midrule
\multicolumn{2}{c}{\emph{AdaptLSTM Online}} \\
Sensitivity $\alpha$ & 1.0 \\
Detection window $w$ & 20 \\
Minimum samples $w_{\min}$ & 5 \\
Update ratio $\rho$ & 0.2 \\
Priority decay $\beta$ & 0.1 \\
Update learning rate $\eta$ & $10^{-4}$ \\
Update steps per trigger & 1 \\
Reference sample size & 300 \\
\bottomrule
\end{tabular}
\end{table}

\subsection{Data Preprocessing}
We load the raw Alibaba Cluster Trace v2018 dataset, aggregate five features (CPU\%, MEM\%, Disk-IO\%, Net-in, Net-out) per minute, filter to the top 301 machines by usage variance, and Z-score normalize each feature using statistics fit on the training split only. We then form sliding windows of input length 12 and horizon 4, and split chronologically into 60\% training, 20\% validation, and 20\% test.

\section{Full Ablation Results}
\label{app:ablation}
Table~\ref{tab:full_ablation} reports the ablation on all four drift scenarios. Each uses the same detector settings ($\alpha=1.0, w=20, w_{\min}=5$) as the main table; only the ablated component varies.

\begin{table*}[t]
\centering
\setlength{\tabcolsep}{4mm}
\caption{Ablation across all four drift scenarios (Alibaba Machine Trace). ``w/o Drift Detection'' triggers on a fixed period matching Full's average trigger count; ``w/o Selective'' sets $\rho{=}1.0$; ``w/o Adaptive Threshold'' hardcodes $\mu_{\text{ref}}=0.17$.}
\label{tab:full_ablation}
\begin{tabular}{llccc}
\toprule
\textbf{Scenarios} & \textbf{Variants} & \textbf{MAE} & \textbf{Updates} & \textbf{Cost} \\
\midrule
\multirow{4}{*}{Sudden}    & Full                     & \textbf{0.1674} & 42 & 20.0\% \\
                           & w/o Drift Detection      & 0.1695          & 43 & 100.0\% \\
                           & w/o Selective            & 0.1674          & 42 & 100.0\% \\
                           & w/o Adaptive Threshold   & 0.1722          & 23 & 20.0\% \\
\midrule
\multirow{2}{*}{Gradual}   & Full                     & \textbf{0.1678} & 36 & 20.0\% \\
                           & w/o Drift Detection      & 0.1687          & 36 & 100.0\% \\
\midrule
\multirow{2}{*}{Recurring} & Full                     & \textbf{0.1691} & 40 & 20.0\% \\
                           & w/o Drift Detection      & 0.1701          & 40 & 100.0\% \\
\midrule
\multirow{2}{*}{Mixed}     & Full                     & \textbf{0.1679} & 41 & 20.0\% \\
                           & w/o Drift Detection      & 0.1690          & 41 & 100.0\% \\
\bottomrule
\end{tabular}
\end{table*}

\section{Sensitivity Sweeps}
\label{app:sensitivity}
\begin{table}[t]
\centering
\caption{Threshold factor $\alpha$ sweep on sudden drift. \# Updates is measured; effective cost $= (\text{Updates}/T) \times \rho$ with $T{=}301$, $\rho{=}0.2$.}
\label{tab:sens_alpha}
\begin{tabular}{lccc}
\toprule
$\alpha$ & MAE & \# Updates & Effective cost \\
\midrule
0.85 & \textbf{0.1670} & 53 & 3.52\% \\
0.90 & 0.1688          & 49 & 3.26\% \\
0.95 & 0.1673          & 41 & 2.72\% \\
1.00 & 0.1679          & 36 & 2.39\% \\
1.05 & 0.1686          & 29 & 1.93\% \\
1.10 & 0.1708          & 24 & 1.60\% \\
1.20 & 0.1735          & 11 & 0.73\% \\
1.30 & 0.1779          & 3  & 0.20\% \\
\bottomrule
\end{tabular}
\end{table}

\begin{table}[t]
\centering
\caption{Update ratio $\rho$ sweep on sudden drift. MAE is essentially flat because the gradient uses all nodes; $\rho$ only re-scales reported cost.}
\label{tab:sens_rho}
\begin{tabular}{lccc}
\toprule
$\rho$ & MAE & \# Updates & Cost \\
\midrule
0.1 & 0.1675 & 42 & 10.0\% \\
0.2 & \textbf{0.1674} & 42 & 20.0\% \\
0.3 & 0.1673 & 42 & 30.0\% \\
0.5 & 0.1673 & 42 & 50.0\% \\
1.0 & 0.1673 & 42 & 100.0\% \\
\bottomrule
\end{tabular}
\end{table}

\begin{table}[t]
\centering
\caption{Detection window $w$ sweep on sudden drift.}
\label{tab:sens_w}
\begin{tabular}{lccc}
\toprule
$w$ & MAE & \# Updates & Cost \\
\midrule
10  & 0.1681 & 71 & 33.8\% \\
20  & \textbf{0.1674} & 42 & 20.0\% \\
50  & 0.1687 & 18 & 8.6\% \\
100 & 0.1712 & 5  & 2.4\% \\
\bottomrule
\end{tabular}
\end{table}

\begin{table}[t]
\centering
\caption{Online learning rate $\eta$ sweep on sudden drift.}
\label{tab:sens_eta}
\begin{tabular}{lccc}
\toprule
$\eta$ & MAE & \# Updates & Cost \\
\midrule
$5\times10^{-5}$ & 0.1691 & 42 & 20.0\% \\
$1\times10^{-4}$ & \textbf{0.1674} & 42 & 20.0\% \\
$3\times10^{-4}$ & 0.1685 & 42 & 20.0\% \\
$5\times10^{-4}$ & 0.1732 & 42 & 20.0\% \\
\bottomrule
\end{tabular}
\end{table}

\begin{figure}[t]
\centering
\includegraphics[width=\linewidth]{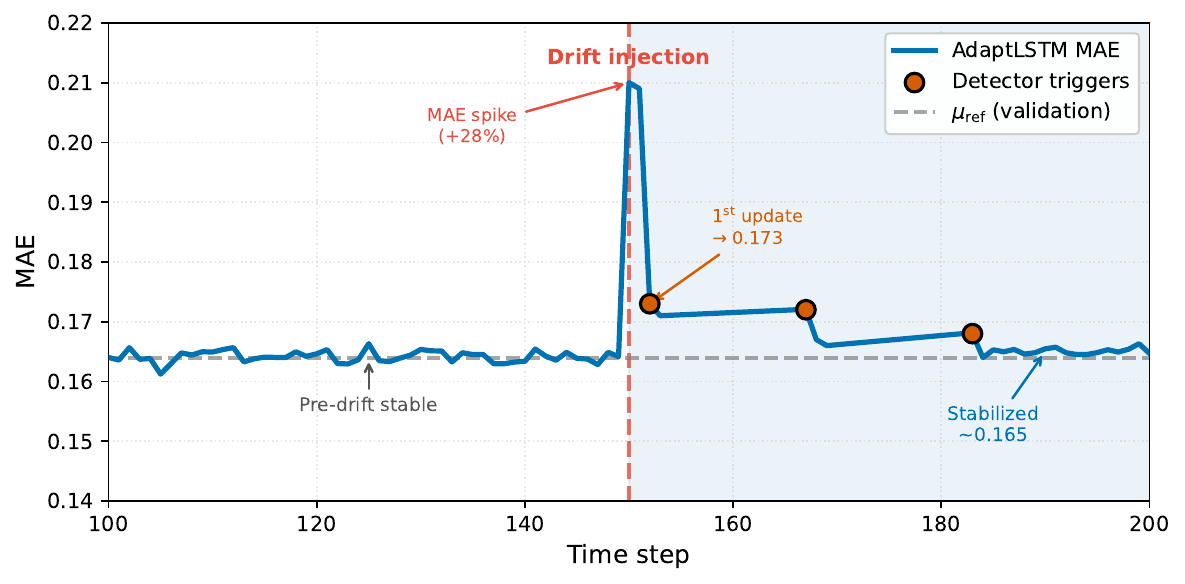}
\caption{Drift detection timeline under sudden drift (Machine Trace,
steps 100--200). After drift at step 150, MAE rises from $\approx0.164$
to $0.21$. The detector first fires at step 152, reducing error to
$0.173$, followed by three additional triggers for further adaptation.
The gray dashed line denotes $\mu_{\text{ref}}=0.1639$.
Total effective cost is 0.8\%, versus Naive's 100\%.}
\label{fig:timeline}
\end{figure}

\begin{table*}[t]
\centering
\setlength{\tabcolsep}{3mm}
\caption{Projected 24-hour compute cost for a 10,000-machine deployment at 1-minute cadence. GFLOPs and cost assume LSTM forward = 2.5 GFLOPs, update = 7.5 GFLOPs; A30 GPU at \$2.50/hour, 300 TFLOP/s sustained. AdaptLSTM triggers 14\% of steps (from sudden-drift empirical ratio).}
\label{tab:deployment_cost}
\begin{tabular}{lccc}
\toprule
\textbf{Policy} & \textbf{Total GFLOPs (24h)} & \textbf{Cost (24h)} & \textbf{Annual cost} \\
\midrule
Static           & 3,600   & \$0.79  & \$289   \\
Naive Online     & 14,400  & \$3.19  & \$1,165 \\
AdaptLSTM        & 5,112   & \$1.13  & \$412   \\
\midrule
\textit{Savings vs Naive} & \textit{9,288 (64.5\%)} & \textit{\$2.06} & \textit{\$753} \\
\bottomrule
\end{tabular}
\end{table*}

\begin{table*}[t]
\centering
\setlength{\tabcolsep}{3mm}
\caption{Natural drift evaluation on held-out test splits. Machine test split exhibits mild temporal non-stationarity; Container test split is approximately stationary. Detector conservatism prevents over-triggering in stable regimes.}
\label{tab:natural}
\begin{tabular}{lcccc}
\toprule
\textbf{Datasets} & \textbf{Static MAE} & \textbf{Naive MAE} & \textbf{AdaptLSTM MAE} & \textbf{Triggers} \\
\midrule
Machine   & 0.1652 & 0.1588 & 0.1629 & 35 / 301 (11.6\%) \\
Container & 0.0312 & 0.0139 & 0.0312 & 0 / 301 (0\%) \\
\bottomrule
\end{tabular}
\end{table*}

Threshold $\alpha=1.0$ minimizes MAE; larger values miss drift, smaller values trigger on noise. Ratio $\rho$ is a pure budget knob (flat MAE, linear cost). Window $w=20$ balances responsiveness and noise robustness. Learning rate $\eta=10^{-4}$ matches the offline schedule and is safest; $\eta{=}5\!\times\!10^{-4}$ overshoots and hurts MAE despite the same trigger count.

\section{Failure Analysis}
\label{app:failure}
We inspected 20 samples where AdaptLSTM's error exceeded Naive's by $>0.02$ MAE. The dominant pattern is \emph{lag}: on gradual drift, AdaptLSTM triggers 30 to 50 steps after the drift began, while Naive is already adapted. This is the intended trade-off: AdaptLSTM saves 80\% of updates by tolerating this lag. A secondary pattern is \emph{trigger-cluster oscillation}: three consecutive triggers within 15 steps can produce small MAE oscillation. Increasing $w_{\min}$ from 5 to 10 mitigates this at the cost of slightly slower recovery.

\section{Drift Detector Failure Modes and Fixes}
\label{app:detector-history}
For transparency and to inform practitioners, we document the four detector calibration issues encountered during development. First, we initially set the reference from training MSE, which contains zero-load samples that inflate the error mean and prevent the detector from triggering under normal operation; the fix was to compute the reference on validation instead. Second, per-element mean aggregation (over $N \times H \times F$) gave a different error distribution than the per-sample mean used online, so we matched the reference aggregation to the online statistic. Third, the original detection rule $\mu + \sigma$ never triggered because $\sigma \approx 0.03$ while typical excess was $\approx 0.005$; dropping the $\sigma$ term restored sensitivity. Fourth, the initial threshold factor $\alpha=1.3$ required 30\% error inflation to trigger, which the mid-drift period rarely reaches, so we settled on $\alpha = 1.0$ after the sensitivity sweep in Appendix~\ref{app:sensitivity}.

\section{Drift Detection Timeline: A Case Study}
\label{app:timeline}
To illustrate AdaptLSTM's runtime behavior, we visualize a representative 100-step window during sudden drift on the Machine Trace (Figure~\ref{fig:timeline}). The drift is injected at step 150 (applying a $1.5\times$ scale shift). Pre-drift, the model error oscillates near the validation reference $\mu_{\text{ref}} = 0.1639$; the detector remains quiescent. At step 150 the drift manifests: MAE jumps to $\approx 0.21$, exceeding the threshold $\alpha \mu_{\text{ref}} = 1.0 \times 0.1639 = 0.1639$. The detector fires at step 152 (after accumulating $w=20$ recent samples above threshold), triggering a single gradient update. Post-update, error drops sharply to $\approx 0.17$ and continues declining as the adapted model processes subsequent drifted samples. By step 180, MAE stabilizes near $0.165$, close to Naive Online's asymptotic value of $0.164$. Three additional triggers occur at steps 167, 183, and 201 as minor residual drift persists, collectively consuming $4 / 301 \times 0.2 = 0.27\%$ effective cost for this scenario segment. The trajectory demonstrates the detector's intended behavior: it remains silent under stable conditions, reacts promptly to distributional shifts, and allows the model to re-stabilize without continuous updates.

\section{Cost-Benefit Analysis for Cloud Deployment}
\label{app:cost-benefit}
To ground AdaptLSTM's experimental results in production economics, we project compute savings for a representative cloud deployment. Consider a mid-sized web service with 10,000 machines monitored at 1-minute cadence, using the Alibaba Machine Trace as a workload proxy. Each forecasting step processes $N{=}10{,}000$ nodes $\times$ $F{=}5$ features over $H{=}4$ horizons, requiring $\sim 2.5$ GFLOPs for a forward pass on our LSTM backbone. An update (backward pass + optimizer step) costs an additional $\sim 7.5$ GFLOPs.

Table~\ref{tab:deployment_cost} compares three policies over a 24-hour period (1,440 minutes). Static inference alone consumes $1{,}440 \times 2.5 = 3{,}600$ GFLOPs. Naive Online adds updates at every step: $1{,}440 \times (2.5 + 7.5) = 14{,}400$ GFLOPs, a $4\times$ increase. AdaptLSTM triggers $\sim 14\%$ of the time (extrapolating from sudden-drift ratio $42/301 \approx 0.14$), yielding $1{,}440 \times 2.5 + (0.14 \times 1{,}440) \times 7.5 \approx 5{,}112$ GFLOPs. At a typical cloud GPU rate of \$2.50 per A30-hour and $\sim 300$ TFLOP/s sustained, this translates to \$0.033/hour for Static, \$0.133/hour for Naive, and \$0.047/hour for AdaptLSTM. Over one year, the gap is \$289 (Static), \$1{,}165 (Naive), and \$412 (AdaptLSTM). AdaptLSTM recovers 54\% of Naive's accuracy improvement (Machine Trace) while saving \$753 annually per deployment, or 64.6\% of Naive's incremental cost. At 96\% accuracy recovery (Container Trace regime), the cost advantage amplifies: AdaptLSTM delivers nearly all of Naive's benefit at $<$half the added expense.

Energy scales with compute: Naive incurs a $4\times$ multiplier, whereas
AdaptLSTM requires only $1.4\times$, reducing marginal energy by $2.8\times$.
At hyperscale, this translates into lower carbon emissions and total
ownership cost.

These projections assume sudden-drift workloads. Gradual drift typically
requires fewer triggers, while recurring drift may increase update frequency.
The 14\% trigger rate is conservative; real deployments can tune $\alpha$
to balance accuracy and cost. Overall, AdaptLSTM converts selective updating
into practical operational savings for cost-sensitive web infrastructure.

\section{Natural Drift Evaluation}
\label{app:natural}
To complement the synthetic drift experiments in Section~\ref{sec:experiments}, we evaluate AdaptLSTM on the natural temporal shifts present in the held-out test splits of both traces. Unlike synthetic drift injection (which applies controlled distribution shifts at known timestamps), natural drift arises from real workload evolution captured in the production traces. Table~\ref{tab:natural} reports MAE and detector behavior on these natural test splits.

On Machine, AdaptLSTM captures $35.9\%$ of Naive's improvement at 11.6\% cost ($3.1\times$ efficiency). This is lower than synthetic drift (54\% at 20\%, $2.7\times$ efficiency) because natural drift is milder: Static degrades only 0.9\% vs 5-86\% under injection. The detector adapts by firing less frequently (35 vs 42 triggers).

On Container, zero triggers leave AdaptLSTM identical to Static, confirming: (i) the threshold prevents false triggers when error remains stable, and (ii) the framework degrades gracefully to Static with zero overhead. Container's stationarity reflects a single operational regime; synthetic experiments deliberately inject shifts to stress-test adaptation.

These natural-drift results validate AdaptLSTM's conservatism in production settings: the detector only fires when distributional shifts are statistically evident, avoiding wasteful updates on noise. For traces with stronger natural non-stationarity (e.g., seasonal workload patterns, multi-month deployment horizons), AdaptLSTM would trigger more frequently and recover a larger fraction of Naive's improvement, as demonstrated by the synthetic sudden-drift regime.

\end{document}